\documentclass{article}

\PassOptionsToPackage{numbers, compress}{natbib}
\usepackage[main,preprint]{neurips_2026}

\usepackage[utf8]{inputenc}
\usepackage[T1]{fontenc}
\usepackage{xcolor}
\definecolor{linkblue}{rgb}{0.12,0.30,0.65}
\definecolor{orange}{rgb}{1.0,0.5,0.0}
\definecolor{stdgray}{gray}{0.45}
\DeclareRobustCommand{\std}[1]{{\textcolor{stdgray}{\smash{\raisebox{0.45ex}{\normalfont\scriptsize$\pm$#1}}}}}
\usepackage[pagebackref,breaklinks,colorlinks,citecolor=linkblue]{hyperref}
\hypersetup{pdftitle={FARE: Forensic Acceptance Region Estimation for Catching Bait-and-Switch Image Generators}, pdfauthor={Kai Yao and Marc Juarez}}
\usepackage{url}
\usepackage{graphicx}
\usepackage{amsmath,amssymb,amsfonts}
\usepackage{booktabs}
\usepackage{multirow}
\usepackage{array}
\usepackage{tabularx}
\usepackage{algorithm}
\usepackage{algpseudocode}
\usepackage{nicefrac}
\usepackage{microtype}
\usepackage{placeins}
\usepackage{caption}

\usepackage{siunitx}
\usepackage{etoolbox}
\robustify\bfseries

\newcolumntype{L}[1]{>{\raggedright\arraybackslash}p{#1}}
\newcolumntype{C}[1]{>{\centering\arraybackslash}p{#1}}
\newcolumntype{Y}{>{\raggedright\arraybackslash}X}
\newcommand{\R}{\mathbb{R}}
\newcommand{\I}{\mathbb{I}}

\newcommand{\argmin}{\mathrm{arg\,min}}

\providecommand{\keywords}[1]{}

\title{FARE: Forensic Acceptance Region Estimation for Catching Bait-and-Switch Image Generators}
\author{%
  Kai Yao\\
  School of Informatics\\
  The University of Edinburgh\\
  \texttt{kai.yao@ed.ac.uk}
  \And
  Marc Juarez\\
  School of Informatics\\
  The University of Edinburgh\\
  \texttt{marc.juarez@ed.ac.uk}
}

\begin{document}

\maketitle

\begin{center}
\small
This work has been accepted for publication in the proceedings of The 40th Annual Conference on Neural Information Processing Systems (NeurIPS 2026).
\end{center}

\begin{abstract}
Modern AI image generators are increasingly deployed as opaque APIs, where customers can query the deployed service, but cannot inspect model weights or architecture. This creates a practical challenge: a provider may pass governance certification with one generator and later silently switch to a cheaper and lower-quality one for deployment, compromising public trust or even safety in high-stakes domains. 
We study \emph{integrity auditing} at deployment time and propose \textbf{FARE} (Forensic Acceptance Region Estimation).
A certified generator is enrolled by training FARE on images sampled from that generator.
After deployment, FARE can determine whether a generated image is consistent with the enrolled generator---using \emph{only} that image.
FARE's features are based on image generator-specific artifacts that have been proposed for forensic applications. FARE amplifies these features during training by finding hard samples that tighten the acceptance region and increase sensitivity to subtle changes in the certified generator.
Across generator swaps, including substitutions with similar model versions and model variants, FARE is effective at detecting swaps, consistently outperforming existing baselines at strict operating points, and remains effective under the exact-model and decision-only attacks evaluated in this work.

\keywords{Trustworthy ML \and model fingerprinting \and model provenance \and auditing \and image forensics \and anomaly detection \and adversarial robustness}
\end{abstract}


\section{Introduction}
\label{sec:intro}

Generative AI is not only widely used by the public, but it is also increasingly considered in safety-critical domains, such as defense~\cite{genai_defense} and healthcare~\cite{genai_medical}. Regulations are emerging to prevent harmful generators from undermining public trust, causing societal harm, or creating security risks. For example, the EU's AI Act establishes conformity-assessment requirements for high-risk AI systems~\cite{AIAct2023}. This requirement raises a key question: \emph{how can external parties, such as auditors or customers, verify that an output from a deployed service truly came from the certified generator?} Even after passing an audit, a provider may replace the certified generator with a cheaper or flawed alternative. This is especially relevant because commercial image generators are often accessed through remote APIs (e.g., \cite{openai2023dalle3,adobe2023firefly}), where external parties can query the service and observe outputs but typically cannot inspect model weights, training data, or other ML pipeline components.

This paper addresses this question as an \emph{integrity auditing} problem using only generator outputs. We formalize it in two phases: \emph{enrollment} and \emph{verification}. During \emph{enrollment}, an auditor certifies a generator under an NDA-like legal agreement and enrolls it into FARE by sampling images. This phase follows governance certification, assumes provider collaboration, and is codified in a \emph{contract} specifying the model checkpoint and inference pipeline settings. Models deviating from the contract, e.g., lower-quality models, are non-certified. During \emph{verification}, the auditor receives only a queried image from the deployed generator and decides whether it is consistent with the certified generator. Here, \emph{output-only} means enrollment and calibration use only images sampled from the certified generator: no auxiliary natural-image negatives, no images from other generators, and no access to generator weights. We focus on the harder \emph{single-image verification} setting over batch verification because deployments often require high-confidence decisions on individual queries, and adversaries could otherwise dilute detection signals by mixing certified and non-certified outputs in a batch.

Why is this setting difficult for an auditor? First, many economically motivated changes are subtle: providers may deploy a closely related checkpoint or lightly modified pipeline whose images look similar to end users. Second, an output-only verifier must rely solely on generator-specific artifacts inferred from the image itself~\cite{marra2018gansfingerprints}. Third, a malicious provider can adapt its deployment to evade a fixed detector via adversarial manipulations~\cite{madry2017towards}.

To address these challenges, we introduce \textbf{FARE (Forensic Acceptance Region Estimation)}, a verifier trained using \emph{only} images from the certified generator. FARE learns an acceptance region for certified outputs in a feature space derived from image artifacts, motivated by forensic evidence that generator artifacts can support source attribution~\cite{marra2018gansfingerprints}. FARE scores image patches locally and uses the most suspicious patch scores to capture spatially sparse artifacts. To meet low-FPR requirements, FARE augments training data with certified-derived ``hard'' examples that reflect contract violations near the decision boundary. At verification, FARE aggregates the most suspicious patch scores into a single image score for the final decision. Our evaluation shows that FARE outperforms compatible baselines across substitution scenarios and remains effective under the evaluated exact-model and decision-only attacks.

In summary, our contributions are:

\textbf{(i) Post-certification auditing under a contract.} We formalize output-only verification of an image generation deployment after certification, where the audit contract covers the model checkpoint and inference pipeline.

\textbf{(ii) FARE: an output-only verifier.} We present a protocol-compatible method that learns a tight acceptance region from only certified outputs using ``forensic'' features, local scoring, and certified-derived hard examples.

\textbf{(iii) Strict-FPR evaluation and batch auditing.} We evaluate model swaps, model-version swaps, cost-motivated inference variants, and exact-model and decision-only attacks at strict operating points. FARE performs strongly in the primary single-image setting at 1\% FPR; at stricter 0.5\% and 0.1\% budgets, we quantify single-image degradation and show that small-batch verification largely recovers performance.

\section{Problem Statement and Threat Model}
\label{sec:threat}
We study \emph{post-certification auditing} of black-box image-generation APIs as a security game between a service provider and an auditor. The auditor observes only output images and must decide whether deployment-time outputs remain compliant with what was certified.

\subsection{Two-phase Auditing: Enrollment vs.\ Verification}

We distinguish a model checkpoint \(M\) from an inference pipeline \(\Pi\) (e.g., number of denoising steps, post-processing policy). A typical deployed generator instance is the pair \(G \;=\; (M,\Pi)\), which induces an output distribution \(P_G\).

\textbf{Enrollment phase (\(t_0\)).}
The provider exposes the certified generator \(G_{\mathrm{cert}}=(M_{\mathrm{cert}},\Pi_{\mathrm{cert}})\) for audit. The auditor has black-box query access to \(G_{\mathrm{cert}}\) and collects $N$ certified outputs
\(
\mathcal{D}_{\mathrm{cert}}=\{x_i\}_{i=1}^{N}, \quad x_i \sim P_{G_{\mathrm{cert}}}.
\)
Using only these outputs, the auditor trains a verifier \(V(\cdot)\). We assume enrollment-time access is honest, i.e., all responses are generated by \(G_{\mathrm{cert}}\). An enrollment \emph{contract} specifies the set of allowed deployments \(\mathcal{C}\subseteq\{(M,\Pi)\}\). In this work, we evaluate a strict singleton contract, \(\mathcal{C}=\{G_{\mathrm{cert}}\},\) so any undeclared change to the checkpoint \(M\) or the pipeline \(\Pi\) is treated as non-certified. More permissive contracts could include pre-agreed pipeline changes, but we do not evaluate them. Appendix~\ref{sec:scenarios} instantiates the strict contract for the generator-swap benchmarks.

\textbf{Verification phase (\(t_1\)).}
After deployment, the provider may serve outputs from some deployment \(G_{\mathrm{dep}}=(M_{\mathrm{dep}},\Pi_{\mathrm{dep}})\), inducing distribution \(P_{G_{\mathrm{dep}}}\). Given only a claimed API output image \(x\), the verifier decides whether \(x\) is consistent with the certified deployment, i.e., whether \(x\) is plausibly drawn from \(P_{G_{\mathrm{cert}}}\) under the agreed contract.

\textbf{Auditor's goal.} The auditor trains only on samples from \(G_{\mathrm{cert}}\) and aims to reject images produced by a broad, unseen set of plausible non-certified deployments \(G_{\mathrm{dep}}\notin\mathcal{C}\) (e.g., model swaps \(M_{\mathrm{dep}}\neq M_{\mathrm{cert}}\), and pipeline changes \(\Pi_{\mathrm{dep}}\neq \Pi_{\mathrm{cert}}\)). We do not aim for universal rejection over arbitrary images.

\subsection{Decision Rule and Compliance Operating Point}

Following anomaly detection, we treat ``non-certified'' as the positive class and use a scalar anomaly score $s(\cdot)\in\R$, where larger values are more suspicious. Given an image $x$, the verifier accepts if
\begin{equation}
V(x)=\I[s(x)\le \tau_\alpha],
\qquad
\tau_\alpha=\mathrm{Quantile}_{1-\alpha}\big(\{s(x_i^{\mathrm{cal}})\}_{i=1}^{m}\big).
\label{eq:decision}
\end{equation}
Here $\{x_i^{\mathrm{cal}}\}_{i=1}^{m}$ is an independent held-out certified calibration set and $\alpha$ is the target certified-image FPR. Appendix~\ref{sec:calibration_uncertainty} specifies the order-statistic threshold and its finite-sample uncertainty. We use 1\% as the primary single-image operating point and also study stricter budgets such as 0.5\% and 0.1\%, which better fit real-world model auditing.

\subsection{Threat Model}

\textbf{Goals.}
The provider is considered malicious only after enrollment. To reduce cost, it may replace $G_{\mathrm{cert}}$ with a cheaper, lower-fidelity deployment, e.g., a lower-tier model, aggressive compression, or altered inference settings. Such undeclared changes can alter output behavior and violate the contract. An active adversary seeks \emph{targeted acceptance}: producing images from a non-certified deployment $G_{\mathrm{dep}} \neq G_{\mathrm{cert}}$ that the verifier falsely accepts as certified. This resembles forging forensic features onto an image so it is wrongly attributed to the certified model~\cite{yao2025smudged}. FARE checks whether an output is consistent with the enrollment contract; it does not infer whether a deviation is benign or malicious. We do \emph{not} claim cryptographic unforgeability.

\textbf{Knowledge and Capabilities.}
After enrollment, the provider controls deployment and may mimic the certified distribution while serving substituted outputs. In deployment, it has black-box, decision-only verifier access under a query budget, with no access to verifier internals or gradients. We evaluate this capability with HopSkipJumpAttack (HSJA)~\cite{chen2020hopskipjump}. We also use exact-model white-box PGD as a high-information stress test in which the attacker knows the verifier's parameters and gradients. This additional access exceeds the deployed threat model, but a particular white-box optimizer is not a universal upper bound on other attacks. We do not evaluate surrogate-transfer, generation-pipeline, or cross-detector attacks.

\section{Related Work}
\label{sec:related}

\textbf{Model Provenance Techniques.}
Watermarking and fingerprinting can attribute image provenance to a source model. Watermarking embeds identifiable signatures post-generation (e.g., \cite{gowal2025synthid,tancik2020stegastamp}) or directly during generation (e.g., \cite{wen2023treering,fernandez2023stablesignature,gunn2024undetectable}). However, watermarking assumes a cooperative provider. If that provider controls the embedding mechanism or key, it can apply the certified watermark to outputs from another generator; watermark presence alone therefore does not establish that the certified generator was used. Passive image forensics instead exploits intrinsic traces left by generators as byproducts of generation, including residual artifacts~\cite{marra2018gansfingerprints,yu2019ganfingerprints} and frequency cues~\cite{dzanic2019fourier,frank2020frequency,durall2020upconv}. Although these features are related to FARE's, prior source-attribution studies typically assume a closed world of possible generators, unlike the open-world setting we evaluate here. Open-world attribution methods such as Girish et al.~\cite{girish2021openworldgan} learn from multiple known sources, unlike our single-source enrollment setting. Recent adversarial evaluations also show that fingerprinting is vulnerable to removal and forgery~\cite{yao2025smudged}.

\textbf{One-Class Verification Methods.}
We formulate open-world attribution to a single source generator as one-class classification. OCC-CLIP~\cite{liu2024occclip} studies few-shot one-class origin attribution with prompt-tuned CLIP, but its full prompt-tuning setup builds a target class from source-model images and a non-target class from auxiliary clean/open-domain images, violating our certified-only enrollment protocol. We therefore include a protocol-compatible CLIP baseline, CLIP + OC-SVM, and exclude full prompt-tuned OCC-CLIP from the evaluation. FLIPAD~\cite{laszkiewicz2024flipad} is also incompatible because it requires white-box generator access. Classical one-class models, such as One-Class SVM~\cite{scholkopf2001ocsvm} and Deep SVDD~\cite{ruff2018deepsvdd}, are included as certified-only baselines under the same information constraints. DROCC~\cite{goyal2020drocc} and DOC3~\cite{dhar2021doc3} motivate auxiliary examples for shaping one-class decision boundaries. Rather than using them as fixed off-the-shelf baselines, we adapt them to our pipeline and ablate each resulting mechanism under identical training, calibration, and scoring conditions. To emphasize generator-specific traces over image content, image forensics uses architectural constraints. The Bayar--Stamm layer~\cite{bayar2016constrained} makes convolutional filters to act as high-pass residual filters, highlighting noise-like artifacts rather than scene content. A patch-based formulation also connects to multiple instance learning, where image-level decisions are formed from local instances~\cite{andrews2002mil,ilse2018attentionmil}.

\section{FARE: Forensic Acceptance Region Estimation}
\label{sec:method}

\begin{figure}[htbp!]
    \centering
    \includegraphics[width=1.0\columnwidth]{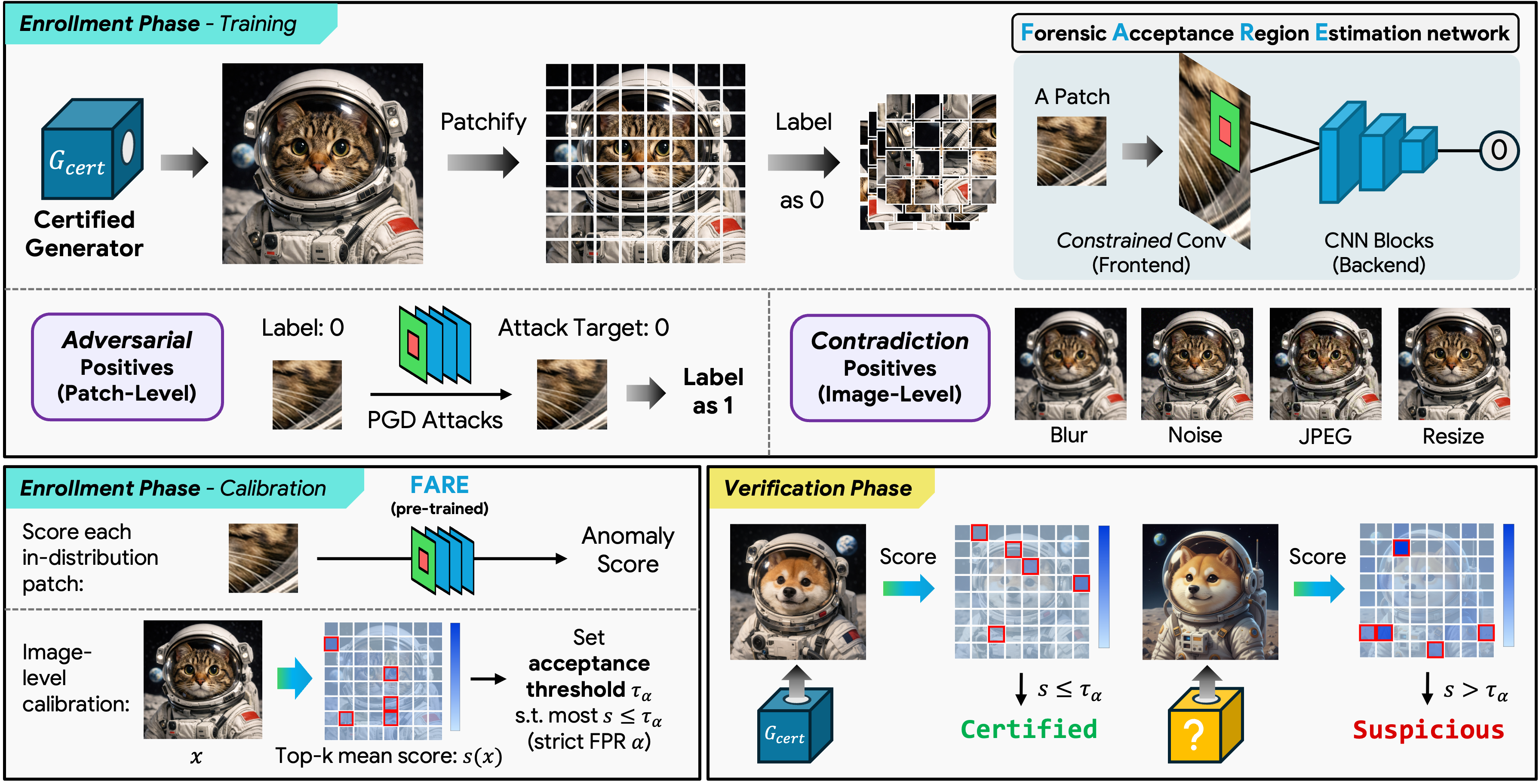}
    \caption{\textbf{FARE.}
    \emph{Enrollment:} using only images from the certified generator, we train a patch verifier in a forensic feature space. Training uses certified patches as negatives and two certified-derived positive sources: (i) near-boundary adversarial positives and (ii) contradiction positives from contract-violating perturbations (then re-tiled into patches). We then calibrate an image-level threshold at target FPR using the same patch tiling and top-$k$ aggregation.
    \emph{Verification:} tile the queried image into patches, score anomalies, aggregate via top-$k$ mean, and accept/reject via Eq.~\eqref{eq:decision}.}
    \label{fig:pipeline}
\end{figure}

\subsection{Overview}
FARE trains an output-only verifier from images produced by the certified generator, then uses deployed API to audit by checking whether a queried image is consistent with the certified acceptance region (Fig.~\ref{fig:pipeline}, Algs.~\ref{alg:fare-train}--\ref{alg:adv-gen}). Our primary goal is reliable \emph{single-image} auditing at the 1\% FPR operating point; when an application demands a lower FPR \emph{and} can afford the operational overhead—namely, collecting multiple outputs from the same deployment and incurring additional query and latency costs—batch verification provides an optional tightening. Implementation-level architecture and training details are deferred to Appendix~\ref{sec:fare_hparams}.

\textbf{Patch tiling.}
Given an RGB image \(x \in \mathbb{R}^{H \times W \times 3}\), we first partition it into a grid of \(d \times d\) square patches to facilitate the extraction of local forensic cues. While patch grids can in general be overlapping, in this work we use a non-overlapping tiling with stride \(d\) and assume \(H\) and \(W\) are divisible by \(d\). For deployments with arbitrary output sizes, the contract should specify a deterministic resize, crop, or padding policy before tiling, and the same preprocessing must be used during enrollment, calibration, and verification.

\textbf{Anomaly scores.}
We train a network on patches $f_\theta:\R^{d\times d\times 3}\!\to\R$ that outputs a logit $a(p) := f_\theta(p)$ where larger $a(p)$ means stronger evidence that $p$ is a non-certified patch. Given $\mathcal{P}(x)=\{p_i\}_{i=1}^{|\mathcal{P}(x)|}$ and anomaly scores $\{a(p_i)\}$, we aggregate with top-$k$ tail averaging. Let $a_{(1)}\ge\cdots\ge a_{(|\mathcal{P}(x)|)}$ denote patch anomaly scores sorted in descending order, the score of an image is then:
\begin{equation}
s(x)
\;=\;
\frac{1}{k}\sum_{j=1}^{k} a_{(j)},
\qquad 1\le k\le |\mathcal{P}(x)|.
\label{eq:image-score}
\end{equation}
Top-$k$ averaging is a compromise between full-image averaging, which can dilute spatially sparse forensic artifacts, and max pooling, which is overly sensitive to a single noisy patch at low FPR. Appendix~\ref{sec:ablation}, Table~\ref{tab:ablation_k} ablates this choice. Finally, we accept/reject with the calibrated rule in Eq.~\eqref{eq:decision}.

\subsection{Enrollment ($t_0$): Learning the Forensic Acceptance Region}
\textbf{FARE network with a constrained convolutional layer.}
The first layer of $f_\theta$ is a Bayar--Stamm constrained convolution \cite{bayar2016constrained} applied per RGB channel. For each channel-specific kernel $W^{(c)}\in\R^{m\times m}$ with center index $(u_0,v_0)$, we enforce $W^{(c)}_{u_0,v_0}=-1, \sum_{(u,v)\neq(u_0,v_0)} W^{(c)}_{u,v}=1,$ so the kernel sum is $0$ and the layer behaves as a learned prediction-error (high-pass) operator. This biases the representation toward residual forensic traces. After each optimizer step, we project $W^{(c)}$ back onto the constraint set. We implement the backend as standard CNN blocks.

\textbf{Warm-up with negative samples only.}
A tight one-class classification boundary can be unstable early in training if hard positives are introduced before the ``certified'' direction is established. We therefore apply a warm-up stage: for the first $T_{\mathrm{warm}}$ iterations, training uses only certified negatives; afterwards we enable adversarial and contradiction positives, which we introduce below.

\textbf{Learning objective.}
Let $\mathrm{BCEWithLogits}(z,y)$ denote binary cross-entropy with logits. Training uses images $x\sim P_{G_{\mathrm{cert}}}$ from $G_{\mathrm{cert}}$ and their patches $p\in\mathcal{P}(x)$.

\textbf{(i) Certified negatives: patches from $G_{\mathrm{cert}}$.}
\begin{equation}
\mathcal{L}_{\mathrm{cert}}
=
\mathbb{E}_{x\sim P_{G_{\mathrm{cert}}}}\;
\frac{1}{|\mathcal{P}(x)|}\sum_{p\in\mathcal{P}(x)}
\Big[\mathrm{BCEWithLogits}\big(f_\theta(p),0\big)\Big].
\label{eq:lcert}
\end{equation}
\textbf{(ii) Near-boundary adversarial positives.}
To tighten the boundary around certified data and harden it locally, for each certified patch $p$, we synthesize a hard positive within an $\ell_2$ shell:
$
\mathcal{S}
=
\big\{\delta\in\R^{d\times d\times 3}\ \big|\ r \le \|\delta\|_2 \le \gamma r \big\}.
$
Here, $r>0$ is the inner radius, $\gamma>1$ controls thickness.
We seek the optimal perturbation $\delta^\star(p)$ that minimizes the certified loss:
\begin{equation}
\delta^\star(p)
=
\underset{\delta\in\mathcal{S}}{\argmin}\;
\mathrm{BCEWithLogits}\!\Big(f_\theta\big(\mathrm{clip}(p+\delta)\big),0\Big),
\qquad
\tilde p=\mathrm{clip}\!\big(p+\delta^\star(p)\big),
\label{eq:adv-gen}
\end{equation}
where $\mathrm{clip}(\cdot)$ clamps to the valid pixel range (e.g., $[0,1]$). We approximate $\delta^\star(p)$ via projected gradient descent (PGD), applying Euclidean projection $\Pi_{\mathcal{S}}$ onto the shell $\mathcal{S}$ at each iteration; reproducibility-oriented pseudocode is provided in Appendix~\ref{sec:fare_algorithms}, Alg.~\ref{alg:adv-gen}. This yields a hard positive $\tilde p$: we push $p$ along the locally most certified-like direction, then inversely train $\tilde p$ as a positive, tightening the decision boundary in the neighborhood of every certified patch:
\begin{equation}
\mathcal{L}_{\mathrm{adv}}
=
\mathbb{E}_{x\sim P_{G_{\mathrm{cert}}}}\;
\frac{1}{|\mathcal{P}(x)|}\sum_{p\in\mathcal{P}(x)}
\Big[\mathrm{BCEWithLogits}\big(f_\theta(\tilde p),1\big)\Big].
\label{eq:ladv}
\end{equation}
\textbf{(iii) Contradiction positives.}
Let $\mathcal{T}$ be a fixed set of contract-violating transforms. We use Gaussian blur, Gaussian noise, JPEG compression, and crop-and-resizing; exact settings are in Appendix~\ref{sec:fare_hparams}, Table~\ref{tab:fare_hparams}. Inspired by learning from contradictions~\cite{dhar2021doc3}, we use transformed certified samples as auxiliary boundary-tightening examples. They are not meant to model the non-certified deployments used at test time, nor the cost-motivated model changes in Table~\ref{tab:costcut}, such as quantization, pruning, or reduced diffusion steps. Instead, they act as certified-derived boundary-tightening perturbations encoding the strict contract assumption: unapproved changes to the output pipeline $\Pi$, including post-processing, should move samples away from the certified acceptance region. If a contract permits JPEG, resizing, or similar post-processing, that operation should be included in $\Pi_{\mathrm{cert}}$ and used consistently for enrollment, calibration, and verification. During training, we apply each $\varphi\in\mathcal{T}$ to form $x^{\mathrm{con}}=\varphi(x)$, and treat re-tiled patches $p^{\mathrm{con}}\in\mathcal{P}(x^{\mathrm{con}})$ as positives:
\begin{equation}
\mathcal{L}_{\mathrm{con}}
=
\mathbb{E}_{x\sim P_{G_{\mathrm{cert}}}}\;
\frac{1}{|\mathcal{T}|}\sum_{\varphi\in\mathcal{T}}\;
\frac{1}{|\mathcal{P}(x^{\mathrm{con}})|}\sum_{p^{\mathrm{con}}\in\mathcal{P}(x^{\mathrm{con}})}
\Big[\mathrm{BCEWithLogits}\big(f_\theta(p^{\mathrm{con}}),1\big)\Big].
\label{eq:lcon}
\end{equation}
The total learning objective to minimize after warm-up then becomes: $\mathcal{L} = \mathcal{L}_{\mathrm{cert}} + \mu\,\mathcal{L}_{\mathrm{adv}} + \lambda\,\mathcal{L}_{\mathrm{con}}.$ Here $\mu,\lambda\ge 0$ trade off adversarial and contradiction terms against the certified negative term.

\textbf{Calibration at target FPR $\alpha$.}
We compute $s(x)$ (Eq.~\eqref{eq:image-score}) on held-out images from $G_{\mathrm{cert}}$ post-training and set $\tau_\alpha$ via Eq.~\eqref{eq:decision}. Calibration and verification use the same patch tiling and aggregation.

\subsection{Verification ($t_1$): Detecting Generator Substitutions}
Given a queried black-box API output image $x$, we (i) tile it into patches $\mathcal{P}(x)$, (ii) compute patch anomalies via a pre-trained FARE network, (iii) aggregate them into an image score $s(x)$ via Eq.~\eqref{eq:image-score}, and (iv) accept/reject via Eq.~\eqref{eq:decision}. The full enrollment/training loop and PGD adversarial-patch routine are given as pseudocode in Appendix~\ref{sec:fare_algorithms}, Algs.~\ref{alg:fare-train}--\ref{alg:adv-gen}.

\textbf{Intuition: Why Does FARE Work?}
FARE’s design is tightly aligned with our threat model: a constrained convolutional layer suppresses unrelated semantic content and highlights forensic traces, patch-level examination and top-$k$ aggregation scoring localize spatially sparse artifacts, and certified-derived ``hard'' positives sharpen the one-class classification boundary against close substitutions and adversarially modified inputs. Together, these choices enhance sensitivity to generator substitutions that induce little semantic shifts. The empirical results in the following sections support this intuition.

\section{Empirical Evaluation}
\label{sec:experiments}


\textbf{Data splits and calibration.}
For each certified generator $G_{\mathrm{cert}}$, we sample disjoint generated-image splits for training, calibration, and verification. Unless stated otherwise, we use 100{,}000 certified training images, 10{,}000 certified calibration images, and 10{,}000 verification images, as detailed in Appendix~\ref{sec:fare_hparams}, Table~\ref{tab:fare_hparams}. The calibration split is used only to set $\tau_\alpha$ following Eq.~\eqref{eq:decision}; FPR/TPR values are computed on the held-out verification split. Appendix~\ref{sec:calibration_uncertainty} quantifies finite-sample uncertainty at the stricter operating points, and Appendix~\ref{sec:sample_efficiency} reports a training-set-size check. All task-specific fitting and calibration use \emph{only} images from $G_{\mathrm{cert}}$, with no auxiliary natural-image negatives, non-certified generator samples, or generator weights introduced during enrollment. Patch-based baselines share the same patch tiling $\mathcal{P}(x)$; their respective score aggregation rules are specified in Appendix~\ref{sec:baseline_impl}.

\textbf{Baselines.}
We compare against single-source verifiers under the same \emph{certified-only} enrollment protocol. Generic one-class baselines include \textbf{Deep SVDD}~\cite{ruff2018deepsvdd}, \textbf{OC-SVM}~\cite{scholkopf2001ocsvm}, and \textbf{CLIP + OC-SVM} using frozen CLIP image embeddings~\cite{radford2021learning}. We also adapt three existing forensic representations as attribution baselines: \textbf{DCT fingerprints} with an OC-SVM~\cite{Giudice21}, and one-class models fitted to frozen \textbf{Forensic Self-Description (FSD)}~\cite{nguyen2025fsd} and \textbf{Neighboring Pixel Relationships (NPR)} features~\cite{tan2024npr}. Task-specific fitting and calibration use only certified-generator outputs, with no natural-image or known non-certified samples introduced during enrollment; this excludes methods such as OCC-CLIP. Finally, \textbf{FARE ablations} remove components within the same patch-verification pipeline and compare against DROCC/DOC3-style boundary tightening~\cite{goyal2020drocc,dhar2021doc3}.

\textbf{Evaluation scenarios.}
All scenarios verify at $t_1$ against non-certified deployments $G_{\mathrm{dep}}\notin\mathcal{C}$ ($\mathcal{C}$ is the contract). We describe each scenario below; detailed case definitions are in Appendix~\ref{sec:scenarios}.

\textbf{(i) Generator swaps (Table~\ref{tab:swaps}).}
We evaluate open-world rejection under cross-family swaps and within-family near-swaps. Cross-family uses different architecture/family pools, while within-family uses the same generator family or version lineage; Appendix~\ref{sec:scenarios}, Table~\ref{tab:scenario_defs} gives exact cases. Cross-family cases include: (i) a pool of 12 generators pre-trained on FFHQ~\cite{Karras2019StyleGAN} (6 GANs, 3 VAEs, 3 diffusion), certifying each model in turn and testing against the remaining 11; and (ii) a Stable Diffusion (SD)~\cite{stable_diffusion} certified generator (SD1.4/1.5/2.0/2.1/3.5)~\cite{sd21,sd35}, evaluated against the diffusion-dominated CommunityForensics pool~\cite{park2025community}. Within-family cases include: (i) StyleGAN2/3 training-configuration variants on FFHQ~\cite{Karras2020StyleGAN2,Karras2021StyleGAN3}; and (ii) SD version swaps among the above checkpoints. For SD experiments, prompts are sourced from DiffusionDB~\cite{diffusiondb}, split disjointly across training, calibration, and verification, and matched between certified and non-certified deployments within each case. We also report a prompt-diversity check for SD in Appendix~\ref{sec:prompt_diversity}, Table~\ref{tab:prompt_diversity}.

\textbf{(ii) Cost-motivated deployment variants (Table~\ref{tab:costcut}).}
We construct same-checkpoint variants motivated by possible resource savings. These include simulated weight compression through quantize--dequantize operations or dense weight pruning, and inference-time changes using fewer diffusion steps, across GAN (StyleGAN3~\cite{Karras2021StyleGAN3}), VAE (VDVAE~\cite{Child2021VDVAE}), and diffusion (NCSN++~\cite{Song2021ScoreSDE}, SD2.1~\cite{sd21}) models. The simulated compression experiments measure detectability, not realized runtime or memory savings; Appendix~\ref{sec:diffusion_efficiency} reports measured resource use for reduced diffusion steps.

\textbf{(iii) Single-image vs.\ batch verification (Table~\ref{tab:batch}).}
Our primary aim is \emph{single-image} auditing at a 1\% FPR. For lower FPR budgets ($0.5\%,0.1\%$), we additionally report \emph{batch verification} by aggregating per-image scores over $n\in\{5,10\}$ images and calibrating batch thresholds on certified batches.

\textbf{Adversarial forgery attacks.}
We evaluate exact-model white-box PGD~\cite{madry2017towards} and decision-only HSJA~\cite{chen2020hopskipjump}. PGD directly minimizes the image score $s(x')$ under an $\ell_\infty$ budget, while HSJA observes only the final binary decision. Appendix~\ref{sec:attack_evaluation} gives both protocols, attack success rates, and their scope.

\textbf{Metrics.}
We report TPR (\%) at fixed FPR $\alpha$, where true positives are detected non-certified outputs and false positives are certified outputs flagged as non-certified. For batch verification with scores $s_i=s(x_i)$, we use $S_{\mathrm{mean}}=\frac{1}{n}\sum_{i=1}^{n}s_i$ and $S_{\mathrm{max}}=\max_i s_i$, rejecting when $S_{\mathrm{mean}}>\tau^{\mathrm{mean}}_{n,\alpha}$ or $S_{\mathrm{max}}>\tau^{\mathrm{max}}_{n,\alpha}$. The thresholds are calibrated on certified batches to satisfy FPR $\alpha$.

\textbf{Implementation notes.}
All generators are sourced from official GitHub or Hugging Face repositories. FARE training and calibration hyperparameters are specified in Appendix~\ref{sec:fare_hparams}, with ablations on key hyperparameters such as patch size $d$ and top-$k$ aggregation in Appendix~\ref{sec:ablation}. The core FARE implementation is publicly available at \url{https://github.com/kaikaiyao/FARE}; Appendix~\ref{sec:repro} describes the release and implementation details.


\section{Results}
\label{sec:results}

\subsection{Swap Detection under Generator Substitution}
\label{subsec:results_swaps}

\begin{table*}[tbp!]
\centering
\small
\resizebox{0.95\textwidth}{!}{%
\begin{tabular}{p{3.9cm}
c
c
c
c}
\toprule
\multirow{3}{*}{\textbf{Detection Method}}
& \multicolumn{2}{c}{\textbf{Cross-Family Swaps}} 
& \multicolumn{2}{c}{\textbf{Within-Family Swaps}} \\
\cmidrule(lr){2-3} \cmidrule(lr){4-5}
& {\textbf{CF-FFHQ}} & {\textbf{CF-Diff}}
& {\textbf{WF-GANTrain}} & {\textbf{WF-SDVer}} \\
\cmidrule(lr){2-2}\cmidrule(lr){3-3}\cmidrule(lr){4-4}\cmidrule(lr){5-5}
& {TPR@1\%FPR} & {TPR@1\%FPR} & {TPR@1\%FPR} & {TPR@1\%FPR} \\
\midrule

Deep SVDD           & 12.46\std{29.05} & 0.00\std{0.00} & 1.23\std{0.62} & 1.21\std{1.06} \\
OC-SVM             & 6.89\std{15.51} & 0.00\std{0.00} & 0.76\std{0.49} & 2.79\std{4.31} \\
CLIP + OC-SVM      & 13.91\std{31.48} & 0.00\std{0.00} & 1.34\std{0.52} & 1.73\std{1.66} \\
DCT Fingerprint + OC-SVM & 11.46\std{26.84} & 0.00\std{0.00} & 0.89\std{0.40} & 3.07\std{4.40} \\
FSD & 8.45\std{13.60} & 18.40\std{3.23} & 1.02\std{0.23} & 11.23\std{3.57} \\
NPR + OC-SVM & 10.92\std{15.15} & 4.60\std{1.12} & 0.77\std{0.31} & 18.32\std{5.19} \\
\midrule

FARE w/o Constraint     & 80.57\std{27.93} & 7.95\std{1.85} & 31.26\std{21.14} & 8.24\std{4.91} \\
FARE w/o Patch          & 77.26\std{30.38} & 10.38\std{3.48} & 11.23\std{10.51} & 7.36\std{4.71} \\
FARE w/o Adv            & 76.60\std{33.05} & 28.77\std{5.18} & 79.34\std{24.26} & 35.33\std{18.23} \\
FARE w/o Contradiction         & 68.06\std{34.01} & 65.78\std{8.87} & 88.92\std{10.25} & 72.08\std{20.75} \\
FARE w/o Adv, Contradiction    & 9.91\std{22.28} & 0.97\std{0.55} & 1.74\std{1.01} & 1.56\std{2.04} \\
\midrule

FARE (Full) & \bfseries 99.43\std{1.19} & \bfseries 99.45\std{0.88} & \bfseries 96.38\std{2.77} & \bfseries 92.56\std{3.06} \\
FARE (Full, under attack) & 99.28\std{1.35} & 99.38\std{0.91} & 95.73\std{2.69} & 92.38\std{3.42} \\
\midrule
Conditional PGD ASR (\%) & 0.15 & 0.07 & 0.67 & 0.19 \\
\bottomrule
\end{tabular}%
}
\caption{%
Single-image swap detection performance under cross-family and within-family swapping scenarios, compared with baselines.
Except for the ASR row, each entry reports TPR@1\%FPR across all evaluated cases within the scenario (details in Appendix~\ref{sec:scenarios}).
Cross-family covers CF-FFHQ (FFHQ-256 model pool) and CF-Diff (diffusion-dominated CommunityForensics pool).
Within-family covers WF-GANTrain (StyleGAN training-configuration variants) and WF-SDVer (Stable Diffusion versioning).
    We also report performance under exact-model white-box PGD against the pretrained FARE network, with $\lVert\delta\rVert_\infty\le0.025$ and $\mathrm{LPIPS}<0.05$. The final row reports conditional ASR: the percentage of initially detected non-certified images changed to accepted under these constraints.
    All task-specific fitting and calibration use only outputs from the certified generator.
}
\label{tab:swaps}
\end{table*}

Table~\ref{tab:swaps} reports FARE's \emph{single-image} swap detection under generator substitution at a strict operating point (TPR@1\%FPR), averaged over all certified--swapped pairs within each scenario. All baselines remain far below FARE in this low-FPR regime. FSD reaches its highest mean TPR on CF-Diff at 18.40\%, while NPR + OC-SVM reaches its highest mean on WF-SDVer at 18.32\%; both remain below 1.1\% on WF-GANTrain. Standard deviations are computed across heterogeneous certified--swapped cases within each scenario; large values indicate that a baseline detects some easy substitutions but fails on many close or diffusion-heavy cases, rather than providing reliable scenario-level performance.

In contrast, FARE achieves near-perfect cross-family detection (99.43\% on CF-FFHQ and 99.45\% on CF-Diff) and remains strong on harder within-family near-swaps (96.38\% on WF-GANTrain and 92.56\% on WF-SDVer). Appendix~\ref{sec:pairwise_results} reports all within-family pairs; the weakest SD pair, SD1.4 $\rightarrow$ SD1.5, reaches 85.85\% TPR. Under the exact-model white-box PGD stress test with $\mathrm{LPIPS}<0.05$, drops are limited (e.g., 99.45\%$\rightarrow$99.38\% on CF-Diff; 92.56\%$\rightarrow$92.38\% on WF-SDVer). Table~\ref{tab:swaps} also reports conditional attack success rate (ASR), the fraction of initially detected non-certified images flipped to accepted while satisfying $\mathrm{LPIPS}<0.05$. On WF-GANTrain, decision-only HSJA reaches 0.14\% ASR at 5{,}000 queries per image, compared with 0.67\% for PGD. PGD reaches 77.64\% ASR without the LPIPS constraint, showing that the low ASR depends on the perceptual criterion. Appendix~\ref{sec:attack_evaluation} gives the full query-budget and LPIPS analyses. These results cover the evaluated objectives and budgets, not black-box attacks in general.

The ablations in Table~\ref{tab:swaps} clarify the role of each component. Removing the constrained layer sharply degrades performance, especially on diffusion-heavy and within-family evaluations (e.g., CF-Diff drops to 7.95\% and WF-SDVer to 8.24\%), supporting the benefit of the constrained residual representation at low FPR. Replacing patch-based scoring with full-image processing also breaks the verifier on the hardest settings (CF-Diff 10.38\%; WF-GANTrain 11.23\%), supporting the benefit of local scoring over full-image processing. Finally, hard positives are essential for a tight boundary: adversarial near-boundary positives and contradiction positives both contribute substantially, and removing both reduces FARE to near chance across scenarios (e.g., CF-Diff 0.97\%; WF-GANTrain 1.74\%). Overall, these results support FARE's design of tightening a content-suppressed one-class acceptance region using certified-derived hard positives. Additional hyperparameter ablations are in Appendix~\ref{sec:ablation}.

\subsection{Detection of Cost-Motivated Deployment Variants}
\label{subsec:results_costcut}

Table~\ref{tab:costcut} evaluates a stricter non-compliance setting than Sec.~\ref{subsec:results_swaps}: instead of swapping to a different generator, the provider deploys \emph{cost-motivated variants of the same checkpoint} (e.g., simulated model compression or fewer inference steps). This is inherently harder because generator-specific traces may persist when the model identity is unchanged; mild modifications can therefore induce only small distribution shifts, making 1\% FPR detection more challenging.

\textbf{Weight compression.}
FARE remains highly sensitive to \emph{aggressive} compression, achieving near-perfect detection under INT4 quantization and heavy pruning across architectures (e.g., 97--99\% TPR on StyleGAN3/VDVAE/NCSN++ and 98.27\% / 98.12\% on SD2.1). In contrast, \emph{milder} compression is harder: INT6 and mild pruning reduce TPR to the low 90s on StyleGAN3/VDVAE/NCSN++ (e.g., 90--93\%) and the high 80s on SD2.1 (86.42\% and 87.36\%). This matches the threat model: stronger compression induces larger shifts, while milder compression remains closer to certified output statistics. We exclude INT8 because this strict-contract benchmark focuses on clearly non-default compression variants, while INT8 is commonly a default and typically preserves fidelity.

\textbf{Diffusion inference-time changes.}
Reducing denoising steps shows a similar severity trend. A moderate reduction ($20\!\rightarrow\!10$) is hardest, especially for SD2.1 (80.86\%), and remains challenging for NCSN++ (88.93\%). A more aggressive reduction ($20\!\rightarrow\!5$) becomes easy to detect (99.12\% on NCSN++ and 98.29\% on SD2.1), indicating that substantial inference-time shortcuts shift outputs away from the certified acceptance region. Overall, FARE reliably flags deployment variants that substantially change the output distribution, while milder same-checkpoint deviations remain harder at strict operating points. This is a limitation of single-image auditing under a strict false-alarm budget: when a deployment change preserves most generator-specific traces, the shift may remain close to the calibrated certified tail.

\begin{table*}[tbp!]
\centering
\small
\resizebox{0.84\textwidth}{!}{%
\begin{tabular}{p{4.2cm}
c
c
c
c}
\toprule
\multirow{2}{*}{\textbf{Deployment Variant}} &
\multicolumn{4}{c}{\textbf{Detection Performance (TPR@1\%FPR)}} \\
\cmidrule(lr){2-5}
& {\textbf{StyleGAN3}} & {\textbf{VDVAE}} & {\textbf{NCSN++}} & {\textbf{SD2.1}} \\
\midrule

\multicolumn{5}{l}{\textit{Weight compression}} \\
\quad Quantization (INT6)      & 92.57\std{0.91} & 91.03\std{1.02} & 90.50\std{1.10} & 86.42\std{1.35} \\
\quad Quantization (INT4)      & 98.89\std{0.34} & 97.14\std{0.48} & 98.39\std{0.29} & 98.27\std{0.31} \\
\quad Mild Pruning           & 92.42\std{0.88} & 93.44\std{0.77} & 91.57\std{0.95} & 87.36\std{1.28} \\
\quad Heavy Pruning           & 98.44\std{0.41} & 99.32\std{0.22} & 97.53\std{0.52} & 98.12\std{0.36} \\
\midrule

\multicolumn{5}{l}{\textit{Inference-time changes}} \\
\quad Diffusion steps: $20\!\rightarrow\!10$   & {--} & {--} & 88.93\std{1.24} & 80.86\std{1.89} \\
\quad Diffusion steps: $20\!\rightarrow\!5$    & {--} & {--} & 99.12\std{0.21} & 98.29\std{0.33} \\
\bottomrule
\end{tabular}%
}
\caption{
    Single-image detection performance of FARE (TPR@1\%FPR) under cost-motivated deployment variants across different generator architectures. For each model, the verifier is trained and calibrated on the default certified deployment, then evaluated on modified variants of the \emph{same} model. Quantization and pruning are simulated weight transformations; reduced diffusion steps change the executed inference schedule.
}
\label{tab:costcut}
\end{table*}

\subsection{Batch Verification at Low FPR}
\label{subsec:results_batch}

\begin{table*}[tbp!]
\centering
\small
\resizebox{0.95\textwidth}{!}{%
\setlength{\tabcolsep}{6pt}%
\begin{tabular}{l c
c c c
c c c}
\toprule
\multirow{2}{*}{\textbf{Rule}} & \multirow{2}{*}{\textbf{$n$}} &
\multicolumn{3}{c}{\textbf{WF-GANTrain (TPR@$\alpha$FPR)}} &
\multicolumn{3}{c}{\textbf{WF-SDVer (TPR@$\alpha$FPR)}} \\
\cmidrule(lr){3-5}\cmidrule(lr){6-8}
& &
{$\alpha=1\%$} & {$\alpha=0.5\%$} & {$\alpha=0.1\%$} &
{$\alpha=1\%$} & {$\alpha=0.5\%$} & {$\alpha=0.1\%$} \\
\midrule
Single-image & 1
& 96.38\std{2.77} & 89.86\std{2.64} & 71.53\std{4.87}
& 92.56\std{3.06} & 85.49\std{3.84} & 62.39\std{5.27} \\
\midrule
Batch (Mean) & 5
& 99.81\std{0.24} & 99.67\std{0.31} & 99.03\std{0.46}
& 99.46\std{0.35} & 99.12\std{0.43} & 98.27\std{0.68} \\
Batch (Mean) & 10
& 99.92\std{0.17} & 99.85\std{0.23} & 99.47\std{0.32}
& 99.73\std{0.26} & 99.41\std{0.36} & 98.84\std{0.49} \\
\midrule
Batch (Max) & 5
& 99.48\std{0.37} & 99.16\std{0.45} & 98.29\std{0.63}
& 98.93\std{0.42} & 98.57\std{0.58} & 97.24\std{0.86} \\
Batch (Max) & 10
& 99.65\std{0.26} & 99.37\std{0.34} & 98.74\std{0.47}
& 99.21\std{0.33} & 98.86\std{0.46} & 97.92\std{0.67} \\
\bottomrule
\end{tabular}%
}
\caption{
Batch verification performance of FARE at given operating points on within-family near-swaps. Entries report mean $\pm$ std TPR@$\alpha$FPR at $\alpha\in\{1\%,0.5\%,0.1\%\}$ for WF-GANTrain (StyleGAN training-configuration variants) and WF-SDVer (Stable Diffusion versioning). 
}
\label{tab:batch}
\end{table*}

Table~\ref{tab:batch} evaluates verification at up to one order of magnitude lower FPR and how much aggregation over multiple outputs from the same deployment recovers performance. On the harder scenarios, FARE is strong for single-image verification at $\alpha{=}1\%$ (96.38\% on WF-GANTrain; 92.56\% on WF-SDVer), but drops at $\alpha{=}0.1\%$ to 71.53\% and 62.39\%, respectively, because the boundary must be calibrated to the certified tail, leaving less margin for subtle swaps. Batching largely closes this gap: with $n{=}5$ at $\alpha{=}0.1\%$, mean aggregation reaches 99.03\% (WF-GANTrain) and 98.27\% (WF-SDVer), while max reaches 98.29\% and 97.24\%; with $n{=}10$, mean reaches 99.47\% and 98.84\%, while max reaches 98.74\% and 97.92\%. The reduced variance relative to single-image verification indicates that aggregation also stabilizes decisions.

Mean aggregation is consistently stronger: averaging reduces certified-batch score variance while accumulating small but consistent substitution-induced shifts. Max aggregation also benefits from batching, but remains slightly weaker at the strictest operating point because fixed-FPR calibration makes the max threshold increasingly conservative as $n$ grows. Overall, Table~\ref{tab:batch} suggests that single-image auditing is effective at $\alpha{=}1\%$ but insufficient at extreme budgets, while small-batch verification (e.g., $n\in\{5,10\}$) achieves high TPR in the evaluated swap scenarios at $\alpha{=}0.1\%$. Applications requiring $\alpha{=}0.1\%$-level false-alarm control should therefore prefer small-batch auditing when repeated queries from the same deployment are available.


\section{Discussion and Conclusion}
\label{sec:discussion}

FARE shows that output-only integrity auditing of black-box image generators is viable under certified-only enrollment. A one-class acceptance region built on forensic residual features, patch-level localization, and certified-derived hard positives achieves near-perfect single-image detection for cross-family substitutions and strong performance on harder within-family near-swaps at 1\% FPR. Table~\ref{tab:swaps} shows that these gains require the full design: removing the constrained convolution, patch scoring, adversarial positives, or contradiction positives each causes substantial degradation. Exact-model white-box PGD and decision-only HSJA have low attack success under the evaluated budgets and the LPIPS constraint of 0.05, although these experiments do not cover every adaptive attack.

Several limitations remain. Mild same-checkpoint variants are hardest because they stay closest to the certified distribution, making small shifts difficult to detect under strict false-alarm budgets. In the evaluated model-swap settings, aggregating five images improves detection at lower FPRs; we have not established the same gain for every compression or inference variant.

We study a strict contract that permits no shift beyond the enrolled deployment. FARE checks whether an image is consistent with that contract; it cannot infer whether an identical output change arose from a benign or malicious pipeline. More permissive contracts could include expected compression, resizing, prompt rewriting, or safety filtering, but we have not evaluated them. We also do not evaluate a versioned commercial API because an opaque service may change its checkpoint without providing an independently verifiable version label. The evaluated generators span GAN, VAE, and diffusion families, but broader commercial deployments remain future work.

Patch-level visualizations and normalized spatial entropy expose the evidence entering FARE's decision and show that the highest-scoring patches are not consistently confined to one fixed image region (Appendix~\ref{sec:patch_evidence}); they do not exclude semantic shortcuts within a patch. Overall, FARE provides evidence for output-only post-certification auditing within this deployment and attack scope. Future directions include broader contracts that tolerate pre-agreed inference variants, hybrid provenance with active watermarking or cryptographic attestation, and extensions to temporal modalities such as video generation, where artifact structures and distribution-shift signals may differ from the spatial setting studied here.

\begin{ack}
We thank our anonymous reviewers for their valuable feedback, which substantially improved the paper. This work was supported by the Edinburgh International Data Facility (EIDF) and the Data-Driven Innovation Programme at the University of Edinburgh. Access to EIDF was facilitated through the University of Edinburgh's Generative AI Laboratory GAIL Fellow scheme. Marc Juarez is a GAIL fellow and the recipient of a Google Research Scholar Award in Security.

The authors declare no competing interests.
\end{ack}

\bibliographystyle{unsrt}
\bibliography{references}

\clearpage
\appendix
\numberwithin{equation}{section}
\numberwithin{figure}{section}
\numberwithin{table}{section}
\numberwithin{algorithm}{section}
\renewcommand{\arraystretch}{1.08}
\setlength{\tabcolsep}{3pt}
\section*{Supplementary Material}
This appendix provides the implementation and experimental details deferred from the main paper. It describes the evaluation scenarios, baseline implementations, FARE hyperparameters and architecture, additional attack and calibration analyses, within-family pairwise results, ablations, sample efficiency, resource use under reduced diffusion steps, patch-level evidence, the Stable Diffusion prompt-diversity check, and reproducibility details.

\section{Detailed Scenario Definition for Generator Swaps}
\label{sec:scenarios}
This section provides detailed definitions for the generator-swap evaluations referenced in Table~\ref{tab:swaps} of the main text. In all experiments, the contract is the strict singleton contract $\mathcal{C}=\{(M_{\mathrm{cert}},\Pi_{\mathrm{cert}})\}$, meaning any undeclared change in either the checkpoint or the inference pipeline is treated as non-certified. Unless otherwise noted, unconditional generators are queried with a fixed latent seed, and conditional text-to-image models are queried with a fixed prompt list. Certified and non-certified deployments are compared under matched seeds or prompts. All images are converted to RGB and, when necessary, resized to the stated evaluation resolution before patch extraction.

This strict contract permits no deployment shift beyond the enrolled checkpoint and pipeline. FARE checks consistency with that contract, not the provider's intent. An image-only auditor cannot distinguish benign and malicious pipelines that produce the same compression, resizing, prompt-rewriting, or safety-filtering artifacts. Such expected changes could be enrolled under a more permissive contract, which we do not evaluate here.

\subsection{Scenario definitions}
Table~\ref{tab:scenario_defs} summarizes the definition of each scenario referenced in the generator-swap evaluations in Table~\ref{tab:swaps}.

For the FFHQ-256 pool, all models are treated as unconditional face generators. For Stable Diffusion experiments, prompts are sampled from DiffusionDB and deduplicated before being split into train, calibration, and verification subsets.

For the CommunityForensics diffusion benchmark, we retain only the diffusion-dominated evaluation subset because it serves as the most relevant open-world stress test for SD-family certification.

The evaluated generators span GAN, VAE, and diffusion families, but we do not evaluate a versioned commercial API. An opaque service may change the served checkpoint without exposing a stable, independently verifiable version label, which would prevent a controlled version-swap experiment.

\begin{table*}[htbp!]
\centering
\caption{Detailed definitions of the evaluation scenarios from Table~\ref{tab:swaps} in the main paper. Abbreviations: SG2 = StyleGAN2, SG3-T = StyleGAN3-T (translation equiv.), SG3-R = StyleGAN3-R (translation and rotation equiv.), SD = Stable Diffusion, ADA = Adaptive Discriminator Augmentation, FFHQU = Unaligned FFHQ dataset}
\label{tab:scenario_defs}
\scriptsize
\begin{tabularx}{\textwidth}{@{}L{0.17\textwidth}L{0.35\textwidth}YC{0.12\textwidth}@{}}
\toprule
Scenario & Certified deployment(s) & Non-certified deployment(s) & Resolution \\
\midrule
CF-FFHQ & SG2, SG3-T, GANformer, StyleSwin, R3GAN, CIPS, VDVAE, VQVAE, NVAE, NCSN++, LDM, ADM & For each certified model, the remaining models in the pool & $256^2$ \\
CF-Diff & SD1.4, SD1.5, SD2.0, SD2.1, SD3.5-Medium & Evaluation images from CommunityForensics & $512^2$ \\
WF-GANTrain (SG2) & SG2-30k, SG2-70k, SG2-140k, SG2-30k-ADA, SG2-70k-ADA, SG2-140k-ADA & For each certified model, the remaining models in the pool & $256^2$ \\
WF-GANTrain (SG3) & SG3-T, SG3-R, SG3-T-FFHQU, SG3-R-FFHQU & For each certified model, the remaining models in the pool & $256^2$ \\
WF-SDVer & SD1.4, SD1.5, SD2.0, SD2.1, SD3.5-Medium & For each certified model, the remaining models in the pool & $512^2$ \\
\bottomrule
\end{tabularx}
\end{table*}

\subsection{Baseline implementation}
\label{sec:baseline_impl}
Table~\ref{tab:baselines} details the one-class and forensic baselines. Their task-specific fitting and threshold calibration use only certified images from the enrolled generator. Patch-based baselines share FARE's patch extraction, with method-specific image-level aggregation reported in the table.

\begin{table*}[htbp!]
\centering
\caption{Implementation details of the one-class and forensic baselines in Table~\ref{tab:swaps}}
\label{tab:baselines}
\footnotesize
\fontsize{8pt}{9pt}\selectfont
\begin{tabularx}{\linewidth}{@{}L{0.24\linewidth}Y@{}}
\toprule
Method & Implementation details \\
\midrule
Deep SVDD & The same convolutional backbone as FARE, but without the constrained front-end and with a $256$-dimensional embedding head. The center is initialized from $5{,}000$ certified patches. Training uses $32\times32$ patches and batch size $16$. The image score is the top-$10$ mean of squared distances to the learned center. \\
OC-SVM (Patch) & A one-class SVM trained on frozen ResNet-50 patch descriptors. Certified $32\times32$ patches are resized to $224\times224$. The RBF kernel uses $\nu=0.01$ and $\gamma=\texttt{scale}$. Image scores are the top-$5$ mean of patch anomaly scores. \\
CLIP + OC-SVM & A one-class SVM trained on frozen CLIP ViT-B/32 image embeddings of a full image. The RBF kernel uses $\nu=0.01$ and $\gamma=\texttt{scale}$. \\
DCT Feature + OC-SVM & A one-class SVM trained on our adapted $63$-dimensional DCT fingerprints, motivated by~\cite{Giudice21}, computed from the mean absolute AC coefficients of each full image. The RBF kernel uses $\nu=0.01$ and $\gamma=\texttt{scale}$. Because this baseline is image-level, no patch aggregation is applied. \\
FSD & A one-class model fitted to the frozen Forensic Self-Description representation~\cite{nguyen2025fsd}. Task-specific fitting and threshold calibration use only certified-generator outputs. \\
NPR + OC-SVM & A one-class SVM fitted to frozen Neighboring Pixel Relationships features~\cite{tan2024npr}. Task-specific fitting and threshold calibration use only certified-generator outputs. \\
\bottomrule
\end{tabularx}
\end{table*}

\section{Hyperparameters of FARE}
\label{sec:fare_hparams}
This section details the hyperparameters used in FARE (Section~\ref{sec:method}). Table~\ref{tab:fare_hparams} lists those for network training and calibration, Table~\ref{tab:arch} describes the network architecture, and Table~\ref{tab:attack_hparams} specifies the exact-model white-box attack.

In line with the Bayar--Stamm design principle, we do not place a non-linearity directly after the constrained layer. All remaining convolutions use GroupNorm and ReLU, and the classifier head outputs a single scalar anomaly logit for each patch.

\begin{table*}[htbp!]
\centering
\caption{FARE network training and calibration hyperparameters. Unless stated otherwise, the same settings are used across all scenarios}
\label{tab:fare_hparams}
\footnotesize
\begin{tabularx}{\textwidth}{@{}L{0.39\textwidth}Y@{}}
\toprule
Hyperparameter & Value \\
\midrule
Num certified training images & 100,000 \\
Num certified calibration images & 10,000 \\
Num verification images & 10,000 \\
Image resolution & $256\times256$ for FFHQ pool (i.e. GAN/VAE/diffusion pool); $512\times512$ for the conditioned SD family \\
Patch size $d$ & $d=32$ for both training and verification \\
Top-$k$ parameter & $k=10$ \\
Optimizer & AdamW \\
Learning rate & $1\times10^{-3}$ \\
Batch size & 32 \\
Max training iterations & 1,000,000 \\
Warm-up iterations $T_{\mathrm{warm}}$ & 2,000 \\
Loss weights $(\mu,\lambda)$ & $(0.1, 0.1)$ \\
Adversarial shell radius $r$ & 0.03 \\
Adversarial shell thickness $\gamma$ & 1.2 \\
PGD steps for adv.\ positives & 5 \\
PGD step size for adv.\ positives & 0.005 before projection onto the $\ell_2$ shell \\
Contradiction transforms $\mathcal{T}$ & Gaussian blur with $\sigma=1.0$ and kernel size $5$; Gaussian noise with std $=2/255$; JPEG quality $=75$; and crop-resize that keeps 90\% of the area and then resizes back \\
\bottomrule
\end{tabularx}
\end{table*}

\begin{table}[htbp!]
\centering
\caption{FARE network architecture}
\label{tab:arch}
\footnotesize
\begin{tabularx}{\linewidth}{@{}L{0.29\linewidth}Y@{}}
\toprule
Stage & Implemented setting \\
\midrule
Constrained conv layer & Bayar--Stamm constrained depthwise convolution, kernel size $5$, eight filters per RGB channel, constraint scale $1.0$ \\
Channel mixing & $1\times1$ conv from $24$ channels to width $64$, followed by ReLU \\
Backbone & Six $3\times3$ convolutional blocks with widths $(64,128,128,256,256,256)$ and strides $(1,2,1,2,1,1)$ \\
Normalization & GroupNorm throughout the backbone \\
Head & Global average pooling, then a linear map from $256$ features to one logit \\
\bottomrule
\end{tabularx}
\end{table}

\begin{table*}[htbp!]
\centering
\caption{Hyperparameters of the exact-model white-box PGD evaluation. Thresholds are kept fixed at the values calibrated on certified images}
\label{tab:attack_hparams}
\footnotesize
\begin{tabularx}{\textwidth}{@{}L{0.31\textwidth}Y@{}}
\toprule
Attack hyperparameter & Value \\
\midrule
Attack type & PGD on image pixels as a white-box stress test against the deployed detector \\
Optimization objective & Minimize $s(x')$, i.e., the image-level detector score after top-$k$ aggregation \\
Perturbation budget $\lVert\delta\rVert_\infty$ & $\epsilon=0.025$ with images normalized to $[0,1]$ \\
Number of PGD steps & 50 with early stopping after successful acceptance \\
Step size & 0.001 \\
Random restarts & 5, retaining the best adversarial example across restarts \\
Attack success criterion & $s(x')\le\tau_{\alpha}$ and $\mathrm{LPIPS}(x',x)<0.05$ \\
Perceptual fidelity check & LPIPS is evaluated on the original and adversarial images \\
\bottomrule
\end{tabularx}
\end{table*}

\section{Algorithmic Pseudocodes}
\label{sec:fare_algorithms}
This section details the pseudocodes referenced from the method section (i.e., Section~\ref{sec:method}). The mathematical objective and verification rule remain in Section~\ref{sec:method}; the algorithms below summarize and specify the training loop and the PGD routine used to generate near-boundary adversarial positives.

\begin{algorithm}[htbp!]
\caption{FARE Enrollment: Training}
\label{alg:fare-train}
\begin{algorithmic}[1]
\Require Certified images $\mathcal{D}_{\mathrm{cert}}$ sampled from marginal $P_{G_{\mathrm{cert}}}$; patch size $d$; transform set $\mathcal{T}$; weights $(\mu,\lambda)$; warm-up iterations $T_{\mathrm{warm}}$; total training iterations $T_{\mathrm{max}}$.
\Ensure Trained verifier network $f_\theta$.
\State Initialize $f_\theta$ with the Bayar--Stamm constrained first layer described in Section~\ref{sec:method}.
\For{$t=1$ \textbf{to} $T_{\mathrm{max}}$}
    \State Sample a minibatch of images $x\in\mathcal{D}_{\mathrm{cert}}$.
    \State Extract patches $\mathcal{P}(x)$ for each $x$.
    \State Compute $\mathcal{L}_{\mathrm{cert}}$ from Eq.~\eqref{eq:lcert}.
    \If{$t \le T_{\mathrm{warm}}$}
        \State Set $\mathcal{L}\gets \mathcal{L}_{\mathrm{cert}}$ \Comment{warm-up: certified negatives only}
    \Else
        \State Compute $\mathcal{L}_{\mathrm{adv}}$ (Eq.~\eqref{eq:ladv}) by generating $\tilde p$ via Alg.~\ref{alg:adv-gen} for each patch.
        \State Compute $\mathcal{L}_{\mathrm{con}}$ (Eq.~\eqref{eq:lcon}) by applying all $\varphi\in\mathcal{T}$ to each $x$.
        \State Set $\mathcal{L}\gets \mathcal{L}_{\mathrm{cert}}+\mu\mathcal{L}_{\mathrm{adv}}+\lambda\mathcal{L}_{\mathrm{con}}$.
    \EndIf
    \State Update $\theta$ by descending $\nabla_\theta \mathcal{L}$.
    \State Project the constrained kernels back onto the Bayar--Stamm constraint set.
\EndFor
\end{algorithmic}
\end{algorithm}

\begin{algorithm}[htbp!]
\caption{Adversarial Patch Generation via PGD}
\label{alg:adv-gen}
\begin{algorithmic}[1]
\Require Certified patch $p$; classifier $f_\theta$; shell $\mathcal{S}$ defined by $(r, \gamma)$; PGD steps $T$; step size $\eta$.
\Ensure Adversarial patch $\tilde p$.
\State Initialize $\tilde p = p$.
\State Initialize random perturbation $\delta$ and apply Euclidean projection $\Pi_{\mathcal{S}}(\delta)$.
\For{$j=1$ \textbf{to} $T$}
    \State Compute gradients: $g \gets \nabla_\delta \mathrm{BCEWithLogits}\big(f_\theta(\mathrm{clip}(\tilde p+\delta)),0\big)$
    \State Update and project: $\delta \gets \Pi_{\mathcal{S}}\!\big(\delta - \eta \cdot g\big)$
\EndFor
\State \Return $\tilde p = \mathrm{clip}(p+\delta)$
\end{algorithmic}
\end{algorithm}

\FloatBarrier

\section{Hyperparameter Ablation Study}
\label{sec:ablation}
This section evaluates the hyperparameters that are key to FARE's performance. Tables~\ref{tab:ablation_d}, \ref{tab:ablation_k}, and \ref{tab:ablation_warmup} present the ablation results for patch size $d$, top-$k$ aggregation, and warm-up length $T_{\mathrm{warm}}$, respectively. For brevity, the WF-GANTrain column reports the average over its SG2 and SG3 variants.

We find that a medium patch size ($d=32$) consistently outperforms both smaller and larger configurations. Regarding aggregation, values of $k$ between $10$ and $50$ yield comparable performance, while smaller or larger values of $k$ perform visibly worse; we fix $k=10$ for all reported results to maintain consistency. Finally, a warm-up period exceeding $1{,}000$ iterations proves beneficial by stabilizing the certified region before adversarial and contradiction positives are introduced, but further increase does not bring significant performance gains. We set $T_{\mathrm{warm}}=2{,}000$ for all experiments.

\begin{table}[htbp]
\centering
\caption{Ablation on patch size $d$. Results are reported as TPR@1\%FPR.}
\label{tab:ablation_d}
\footnotesize
\begin{tabular*}{\linewidth}{@{\extracolsep{\fill}}lcccc@{}}
\toprule
$d$ & CF-FFHQ & CF-Diff & WF-GANTrain & WF-SDVer \\
\midrule
8   & 91.24\std{1.25} & 99.17\std{0.67} & 92.42\std{2.43} & 90.35\std{5.42} \\
32  & \textbf{99.43}\std{1.19} & \textbf{99.45}\std{0.88} & \textbf{96.38}\std{2.77} & \textbf{92.56}\std{3.06} \\
128 & 89.93\std{2.51} & 85.02\std{4.53} & 70.11\std{6.12} & 72.84\std{9.43} \\
\bottomrule
\end{tabular*}
\end{table}

\begin{table}[htbp]
\centering
\caption{Ablation on the top-$k$ image aggregation parameter. Results are reported as TPR@1\%FPR.}
\label{tab:ablation_k}
\footnotesize
\begin{tabular*}{\linewidth}{@{\extracolsep{\fill}}lcccc@{}}
\toprule
$k$ & CF-FFHQ & CF-Diff & WF-GANTrain (avg.) & WF-SDVer \\
\midrule
5   & 98.87\std{1.12} & 96.08\std{2.34} & 94.86\std{1.89} & 88.73\std{3.45} \\
10  & 99.43\std{1.19} & \textbf{99.45}\std{0.88} & \textbf{96.38}\std{2.77} & \textbf{92.56}\std{3.06} \\
50  & \textbf{99.49}\std{0.74} & 99.11\std{1.02} & 95.74\std{1.72} & 91.88\std{2.41} \\
100 & 95.54\std{2.67} & 94.27\std{3.11} & 92.91\std{3.48} & 86.92\std{4.92} \\
\bottomrule
\end{tabular*}
\end{table}

\begin{table}[htbp]
\centering
\caption{Ablation on the certified-only warm-up length $T_{\mathrm{warm}}$. Results are reported as TPR@1\%FPR.}
\label{tab:ablation_warmup}
\footnotesize
\begin{tabular*}{\linewidth}{@{\extracolsep{\fill}}lcccc@{}}
\toprule
$T_{\mathrm{warm}}$ & CF-FFHQ & CF-Diff & WF-GANTrain & WF-SDVer \\
\midrule
0     & 96.26\std{2.14} & 98.71\std{1.05} & 93.54\std{2.88} & 89.92\std{2.12} \\
1,000 & 98.90\std{1.45} & 99.23\std{0.82} & 95.81\std{1.94} & \textbf{92.84}\std{1.56} \\
2,000 & \textbf{99.43}\std{1.19} & 99.45\std{0.88} & \textbf{96.38}\std{2.77} & 92.56\std{3.06} \\
10,000 & 99.40\std{1.02} & \textbf{99.68}\std{0.54} & 96.33\std{1.68} & 92.13\std{2.04} \\
\bottomrule
\end{tabular*}
\end{table}

\section{Additional Evaluation Analyses}
\label{sec:additional_analyses}

\subsection{Attack Evaluation}
\label{sec:attack_evaluation}

We report attack success rate (ASR) as the fraction of initially detected non-certified images that the attack changes to accepted at the fixed calibrated threshold. The exact-model white-box attack uses PGD with an $\ell_\infty$ budget of $\epsilon=0.025$ for pixels in $[0,1]$. An attack succeeds only when it changes the decision and satisfies $\mathrm{LPIPS}(x_{\mathrm{adv}},x)<0.05$. Under this criterion, conditional ASR is 0.15\% on CF-FFHQ, 0.07\% on CF-Diff, 0.67\% on WF-GANTrain, and 0.19\% on WF-SDVer.

The LPIPS condition substantially changes the PGD result. On WF-GANTrain, ASR is 0.05\%, 0.40\%, 0.67\%, 5.53\%, 23.86\%, and 58.10\% at LPIPS thresholds of 0.01, 0.025, 0.05, 0.1, 0.25, and 0.5, respectively; without an LPIPS constraint, it is 77.64\%. The higher success rates therefore require progressively larger changes to the image.

We also evaluate decision-only HopSkipJumpAttack (HSJA)~\cite{chen2020hopskipjump} on WF-GANTrain under the same LPIPS threshold of 0.05. The attacker observes only FARE's binary accept-or-reject decision. ASR is 0.00\% at 100 and 500 queries per image, 0.05\% at 1{,}000 queries, and 0.14\% at 5{,}000 queries, compared with 0.67\% for exact-model PGD in the same scenario. These results cover the evaluated exact-model and decision-only attacks. They do not establish robustness to surrogate-transfer, generation-pipeline, cross-detector, or other adaptive attacks.

\subsection{Finite-Sample Calibration at Strict Operating Points}
\label{sec:calibration_uncertainty}

The calibrated false-positive rate varies across finite calibration samples even when the scorer is fixed. Let $Z_1,\ldots,Z_m$ be independent and identically distributed certified single-image scores with continuous cumulative distribution function $F$, and write their ascending order statistics as $Z_{(1)}\leq\cdots\leq Z_{(m)}$. For an upper-tail rank $r_\alpha$, distinct from the patch-aggregation parameter $k$, set $\tau_\alpha=Z_{(m+1-r_\alpha)}$. The population false-positive rate at this random threshold is $p_T=1-F(\tau_\alpha)$ and follows
\begin{equation}
p_T \sim \mathrm{Beta}(r_\alpha,m+1-r_\alpha).
\end{equation}
This follows because the probability integral transform makes $F(Z_i)$ independent uniform variables, so $F(\tau_\alpha)$ is their $(m+1-r_\alpha)$th order statistic. With $m=10{,}000$, the nominal 0.5\% and 0.1\% targets use $r_\alpha=50$ and $r_\alpha=10$. Their exact central 95\% ranges are $[0.371\%,0.647\%]$ and $[0.048\%,0.171\%]$, and their one-sided 95\% upper bounds are 0.621\% and 0.157\%. Thus, a nominal target does not fix the population FPR for every realized calibration set. These intervals assume continuous scores; tied scores require a specified tie-breaking rule and corresponding analysis. They quantify finite-calibration uncertainty for the fixed single-image scorer and the same certified distribution. They do not include scorer or distribution changes, or automatically provide a guarantee for batch calibration.

\subsection{Pairwise Results for Within-Family Swaps}
\label{sec:pairwise_results}

Tables~\ref{tab:pairwise_sg2}, \ref{tab:pairwise_sg3}, and~\ref{tab:pairwise_sd} report every certified--substituted pair in the within-family scenarios. Rows identify the certified generator and columns identify the substituted generator. The weakest pairs are SG2-70k-ADA $\rightarrow$ SG2-140k-ADA at 89.29\%, SG3-R-FFHQU $\rightarrow$ SG3-R at 95.18\%, and SD1.4 $\rightarrow$ SD1.5 at 85.85\% TPR at 1\% FPR.

\begin{table*}[htbp]
\centering
\scriptsize
\begin{tabular}{lrrrrrr}
\toprule
Certified & 30k & 70k & 140k & 30k-A & 70k-A & 140k-A \\
\midrule
SG2-30k & -- & 97.40 & 98.04 & 96.81 & 98.28 & 98.12 \\
SG2-70k & 98.45 & -- & 91.67 & 98.08 & 93.49 & 93.82 \\
SG2-140k & 99.20 & 92.52 & -- & 96.71 & 93.67 & 93.01 \\
SG2-30k-ADA & 97.72 & 97.14 & 97.54 & -- & 94.44 & 94.78 \\
SG2-70k-ADA & 99.48 & 94.31 & 94.56 & 95.36 & -- & 89.29 \\
SG2-140k-ADA & 99.53 & 94.73 & 93.82 & 95.83 & 90.11 & -- \\
\bottomrule
\end{tabular}
\caption{Pairwise WF-GANTrain results for StyleGAN2 training variants. Entries are TPR@1\%FPR (\%); A denotes ADA training.}
\label{tab:pairwise_sg2}
\end{table*}

\begin{table}[htbp]
\centering
\scriptsize
\begin{tabular}{lrrrr}
\toprule
Certified & SG3-T & SG3-R & SG3-T-FFHQU & SG3-R-FFHQU \\
\midrule
SG3-T & -- & 99.01 & 96.93 & 99.93 \\
SG3-R & 99.83 & -- & 99.93 & 96.11 \\
SG3-T-FFHQU & 96.02 & 99.19 & -- & 98.98 \\
SG3-R-FFHQU & 99.20 & 95.18 & 99.82 & -- \\
\bottomrule
\end{tabular}
\caption{Pairwise WF-GANTrain results for StyleGAN3 variants. Entries are TPR@1\%FPR (\%).}
\label{tab:pairwise_sg3}
\end{table}

\begin{table}[htbp]
\centering
\scriptsize
\begin{tabular}{lrrrrr}
\toprule
Certified & SD1.4 & SD1.5 & SD2.0 & SD2.1 & SD3.5-M \\
\midrule
SD1.4 & -- & 85.85 & 90.29 & 91.02 & 93.32 \\
SD1.5 & 89.53 & -- & 89.58 & 90.54 & 93.96 \\
SD2.0 & 94.06 & 93.97 & -- & 87.63 & 93.40 \\
SD2.1 & 94.01 & 94.12 & 90.87 & -- & 92.88 \\
SD3.5-M & 97.03 & 95.95 & 96.57 & 96.62 & -- \\
\bottomrule
\end{tabular}
\caption{Pairwise WF-SDVer results. Entries are TPR@1\%FPR (\%).}
\label{tab:pairwise_sd}
\end{table}

\FloatBarrier

\subsection{Contradiction Transform Ablation}
\label{sec:contradiction_ablation}

We remove each contradiction transform in turn from the four-transform set used during training. In the CF-FFHQ pool, the StyleGAN2-certified detector reaches 98.84\% average TPR at 1\% FPR with all four transforms. Removing any transform reduces performance, although the effect is not uniform: crop--resize and Gaussian blur have the largest effects in this setting.

\begin{table}[htbp]
\centering
\small
\begin{tabular}{lr}
\toprule
Contradiction transform set & Mean TPR@1\%FPR \\
\midrule
All four transforms & 98.84 \\
Without Gaussian blur & 86.76 \\
Without Gaussian noise & 94.28 \\
Without JPEG compression & 93.90 \\
Without crop--resize & 80.63 \\
\bottomrule
\end{tabular}
\caption{Leave-one-out ablation of the contradiction transforms for a StyleGAN2-certified detector in CF-FFHQ (\%).}
\label{tab:contradiction_leave_one_out}
\end{table}

\subsection{Training-Set Size}
\label{sec:sample_efficiency}

The main experiments use 100{,}000 training images as a consistent protocol across settings, not as a universal minimum. In CF-FFHQ, FARE reaches 99.27\% TPR at 1\% FPR with 1{,}000 training images, compared with 99.43\% using 100{,}000. In WF-SDVer, 10{,}000 training images approach the performance obtained with 100{,}000, while only a few thousand produce a noticeable reduction. The required training-set size may therefore depend on how close the substituted generators are to the certified generator. The calibration set remains fixed at 10{,}000 images throughout this study; calibration-set size is not separately ablated.

\subsection{Resource Use under Fewer Diffusion Steps}
\label{sec:diffusion_efficiency}

Reducing the number of diffusion steps changes the executed computation directly. Table~\ref{tab:diffusion_efficiency} reports the measured latency, estimated GPU board energy, peak memory, and CLIP image--text similarity for the evaluated 20-, 10-, and 5-step configurations. Relative to 20 steps, 10 steps reduce latency by 46.5\% and estimated energy by 42.3\%; 5 steps reduce them by 70.7\% and 68.6\%. Peak memory is unchanged, while lower CLIP similarity indicates weaker prompt alignment. We do not infer monetary savings because hardware and provider pricing differ.

\begin{table}[htbp]
\centering
\footnotesize
\begin{tabular}{rrrrr}
\toprule
Steps & Latency (s/image) & Energy (J/image) & Peak memory (GiB) & CLIP similarity \\
\midrule
20 & 1.311 & 354.9 & 3.10 & 0.2956 \\
10 & 0.701 & 204.9 & 3.10 & 0.2796 \\
5 & 0.384 & 111.4 & 3.10 & 0.2409 \\
\bottomrule
\end{tabular}
\caption{Resource and prompt-alignment measurements for the reduced diffusion-step experiment. Energy is estimated GPU board energy.}
\label{tab:diffusion_efficiency}
\end{table}

The quantization and pruning variants in Table~\ref{tab:costcut} do not provide corresponding resource measurements. Quantization is simulated through quantize--dequantize operations while retaining the original tensor storage and computation. Pruning zeros the specified proportion of smallest-magnitude weights without sparse storage or kernels. These experiments measure detectability after simulated compression, not realized latency, memory, or energy savings.

\subsection{Patch-Level Evidence}
\label{sec:patch_evidence}

\begin{figure*}[htbp]
\centering
\includegraphics[width=\linewidth]{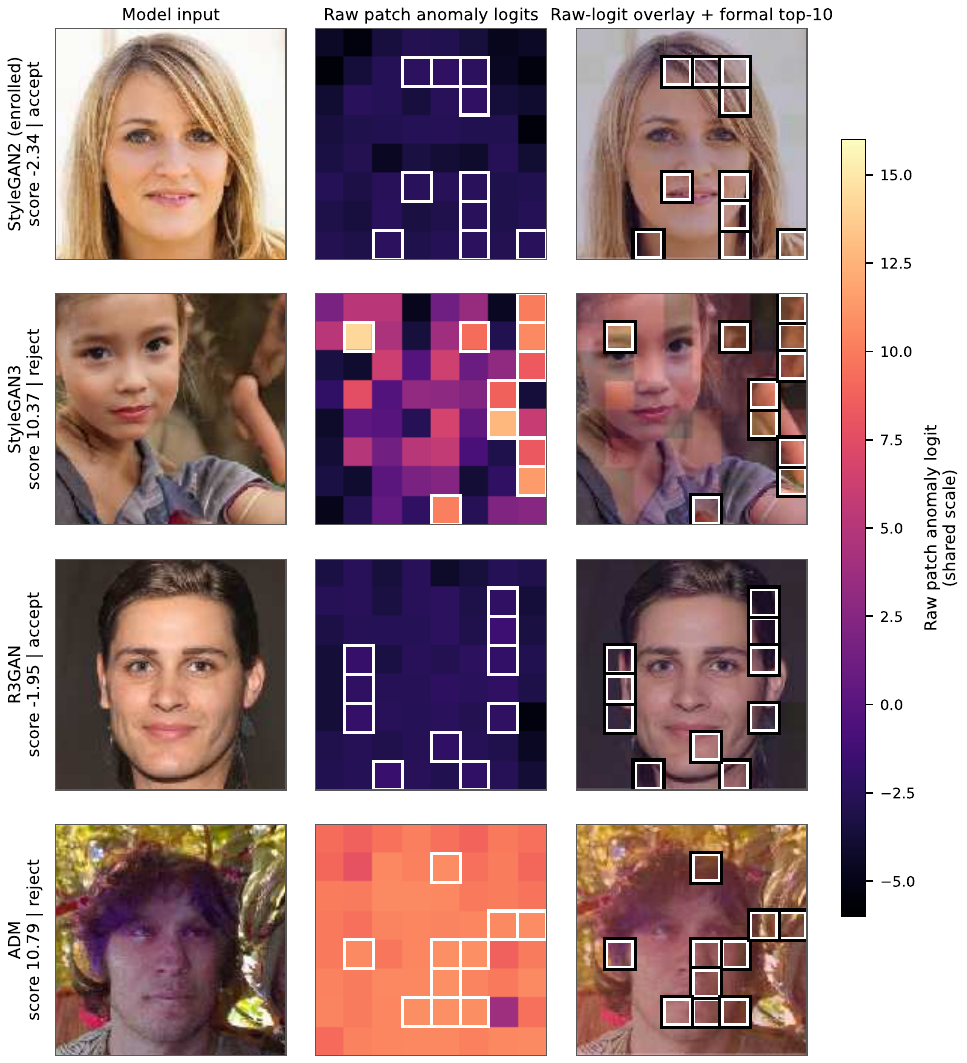}
\caption{Representative FARE patch evidence. Each row shows the model input, raw patch anomaly logits on a shared scale, and the logits overlaid on the input. White boxes mark the ten patches averaged by the image-level decision rule. The examples include the enrolled StyleGAN2 generator and substituted StyleGAN3, R3GAN, and ADM generators. The R3GAN image is accepted despite being a substitute, illustrating a false negative.}
\label{fig:patch_evidence}
\end{figure*}

FARE divides each $256\times256$ image into an $8\times8$ grid of non-overlapping $32\times32$ patches. Its image score averages the ten largest raw patch anomaly logits. Figure~\ref{fig:patch_evidence} shows one representative image from the enrolled StyleGAN2 generator and from each of three substituted generators. The white boxes identify the same ten patches used by the decision rule; the visualization is therefore a direct view of the evidence entering the score rather than a separate post-hoc attribution method.

To test whether these selected patches repeatedly occupy one fixed image region, we separately evaluate 100 images from each generator and count how often every grid cell enters the top ten. Normalized spatial entropy ranges from 0.965 to 0.992 across the four generators, indicating that the selected patches are broadly distributed over the grid. This rules out consistent reliance on one fixed patch location in this experiment. It does not rule out semantic shortcuts within a patch or establish that every selected feature is causally forensic.

\FloatBarrier

\section{Prompt-Diversity Check for Stable Diffusion}
\label{sec:prompt_diversity}
The prompt-diversity check evaluates whether FARE remains effective when the prompt distribution at verification time differs from the distribution used during enrollment and calibration. We draw prompts from PartiPrompts and split them into disjoint buckets based on the dataset's \texttt{Category} field. For each run, training and calibration use prompts from a single category, while verification uses prompts drawn from all remaining categories. Table~\ref{tab:prompt_diversity} reports four such single-category enrollment settings: ``Animal'', ``Artifacts'', ``World-knowledge'', and ``People''. Results are averaged over the Stable Diffusion version-pair substitutions used in the WF-SDVer scenario.

Across the four held-out category splits, FARE reaches 97.74--99.27\% TPR at 1\% FPR. These results show that FARE remains effective under the evaluated category-level prompt shifts. They do not cover more extreme content such as text-heavy images, diagrams, medical-style images, abstract art, or unusual textures, and they do not by themselves establish that the detector is independent of image semantics.

\begin{table}[!ht]
\centering
\caption{Prompt-diversity robustness for Stable Diffusion. Results are reported as TPR@1\%FPR under held-out prompt distributions}
\label{tab:prompt_diversity}
\footnotesize
\begin{tabularx}{\linewidth}{@{}L{0.39\linewidth}L{0.31\linewidth}C{0.18\linewidth}@{}}
\toprule
Train/cal.
prompt split & Verification prompt split & TPR@1\%FPR \\
\midrule
``Animal'' & Other prompts & 97.83\std{2.25} \\
``Artifacts'' & Other prompts & 99.27\std{1.03} \\
``World-knowledge'' & Other prompts & 97.74\std{1.40} \\
``People'' & Other prompts & 98.68\std{1.79} \\
\bottomrule
\end{tabularx}
\end{table}

\section{Reproducibility}
\label{sec:repro}
Our codebase for reproducing FARE is publicly available at \url{https://github.com/kaikaiyao/FARE}. It provides the verifier, training, calibration, scoring and PGD attack utilities, method configurations, unit tests, and a training and scoring example. The README provides installation and data preparation instructions, together with links to the publicly available FFHQ, DiffusionDB, and CommunityForensics resources. Generator choices, data splits, hyperparameters, and evaluation protocols are specified in this paper and appendix.

\begin{table}[!ht]
\centering
\caption{Reproduction details: software, hardware, and code publication}
\label{tab:reproducibility}
\footnotesize
\begin{tabularx}{\linewidth}{@{}L{0.2\linewidth}Y@{}}
\toprule
Item & Value \\
\midrule
Software stack & Python 3.11.10, PyTorch 2.5.1, CUDA 12.1, torchvision 0.20.1, diffusers 0.37.0, transformers 5.3.0, scikit-learn 1.8.0, pandas 3.0.1 \\
GPU & NVIDIA H200 \\
Code release & \url{https://github.com/kaikaiyao/FARE} (core implementation) \\
\bottomrule
\end{tabularx}
\end{table}

\FloatBarrier


\end{document}